\documentclass[letterpaper, 10 pt, conference]{ieeeconf}  

\IEEEoverridecommandlockouts                              

\usepackage{multicol}
\usepackage[bookmarks=true]{hyperref}
\usepackage{graphicx}
\usepackage{amsmath}
\usepackage{amssymb}
\graphicspath{{./images/}}
\usepackage[linesnumbered,ruled,vlined]{algorithm2e} 

\SetCommentSty{mycommfont}
\usepackage{authblk}
\usepackage{xcolor}
\usepackage{comment}
\usepackage{diagbox}
\usepackage{wrapfig}
\usepackage{makecell}
\usepackage{arydshln}

\usepackage{multirow}

\title{\LARGE \bf Modular Adaptive Policy Selection for Multi-Task Imitation Learning through Task Division}

\author{Dafni Antotsiou$^1$, Carlo Ciliberto$^2$ and Tae-Kyun Kim$^{1,3}$
\thanks{$^{1}$ Imperial College London, UK, \{d.antotsiou17, tk.kim\}@imperial.ac.uk}
\thanks{$^{2}$ University College London, UK, c.ciliberto@ucl.ac.uk}
\thanks{$^{3}$ Supported by the Ministry of Land, Infrastructure and Transport of Korea (22CTAP-C163793-02), and the National Research Council of Science and Technology, Korea (CRC 21011).}
}

\makeatletter
\newcommand{\subalign}[1]{%
  \vcenter{%
    \Let@ \restore@math@cr \default@tag
    \baselineskip\fontdimen10 \scriptfont\tw@
    \advance\baselineskip\fontdimen12 \scriptfont\tw@
    \lineskip\thr@@\fontdimen8 \scriptfont\thr@@
    \lineskiplimit\lineskip
    \ialign{\hfil$\m@th\scriptstyle##$&$\m@th\scriptstyle{}##$\hfil\crcr
      #1\crcr
    }%
  }%
}
\makeatother

\begin{document}

\maketitle
\thispagestyle{empty}
\pagestyle{empty}

\begin{abstract}
Deep imitation learning requires many expert demonstrations, which can be hard to obtain, especially when many tasks are involved. However, different tasks often share similarities, so learning them jointly can greatly benefit them and alleviate the need for many demonstrations. But, joint multi-task learning often suffers from negative transfer, sharing information that should be task-specific. In this work, we introduce a method to perform multi-task imitation while allowing for task-specific features. This is done by using proto-policies as modules to divide the tasks into simple sub-behaviours that can be shared. The proto-policies operate in parallel and are adaptively chosen by a selector mechanism that is jointly trained with the modules. Experiments on different sets of tasks show that our method improves upon the accuracy of single agents, task-conditioned and multi-headed multi-task agents, as well as state-of-the-art meta learning agents. We also demonstrate its ability to autonomously divide the tasks into both shared and task-specific sub-behaviours.
\end{abstract}

\section{INTRODUCTION}
In the field of imitation learning (IL), an autonomous agent learns to perform tasks by mimicking demonstrations provided by an expert~\cite{hussein2017imitation}. In control settings, there has been significant work tackling imitation of a single task~\cite{ho2016generative,jena2020augmenting}, as well as generalising around it~\cite{li2017infogail,wang2017robust,hausman2017multi} in recent years. However, learning multiple tasks separately is not always feasible or ideal, which is why the multi-task learning (MTL) paradigm~\cite{zhang2021survey} is used to combine the learning process of many tasks simultaneously. This combination can be challenging, because it not only needs to share commonalities between the tasks so that it performs better than learning them individually (positive transfer), but also avoid sharing unwanted dissimilarities that might hinder training, compared to the individual result (negative transfer)~\cite{caruana1997multitask}. In our work, we aim to tackle this problem by introducing parallel modules that break down the shareable and non-shareable parts of tasks. 
\par
MTL can be particularly useful in deep IL dealing with control problems, since obtaining large expert datasets for robot execution can be challenging and time-consuming~\cite{antotsiou2018task}. The sharing of knowledge and datasets can not only improve performance, but also reduce the network size, compared to using a separate agent for each task~\cite{deisenroth2014multi}. This strategy also reduces the number of hyper-parameters requiring tuning during training, since all tasks are trained at once. MTL is also closely related to meta-learning. The objective of meta-learning in MTL and transfer learning (TL) is to find a common representation for all the tasks, so that agents can quickly converge to new tasks from this common point~\cite{finn2017model,singh2020scalable,yu2018one,nagabandi2018learning}. However, this requires all tasks to be similar in the state space and similar in complexity, since it uses a common base policy and does not account for negative transfer. This can be tackled by either combining opposing gradient updates~\cite{chen2018gradnorm,yu2020gradient}, or by using a network which changes architecture based on the task. One way to achieve dynamic architecture is through the use of adaptive networks, which have multiple optional layers -- either in parallel~\cite{rosenbaum2017routing,ramachandran2018diversity}, sequential~\cite{sun2019adashare,vandenhende2019branched} or both~\cite{purushwalkam2019task} -- and a router that selects which of these layers to activate for each task. This architecture can be particularly beneficial to image problems -- especially given the many different types of layers that can be used as experts~\cite{ramachandran2018diversity,shazeer2017outrageously}. However, control problems do not usually use very deep or convoluted architectures, so the benefits from such methods can be limited. But even cases where similar modules are used as streams, such as~\cite{kendall2018multi}, face additional challenges in control, where the summation of the outputs of all the tasks can lead to erroneous results, since different tasks can have opposing actions.
\begin{figure}[t]
\centering
\includegraphics[width=0.4\textwidth]{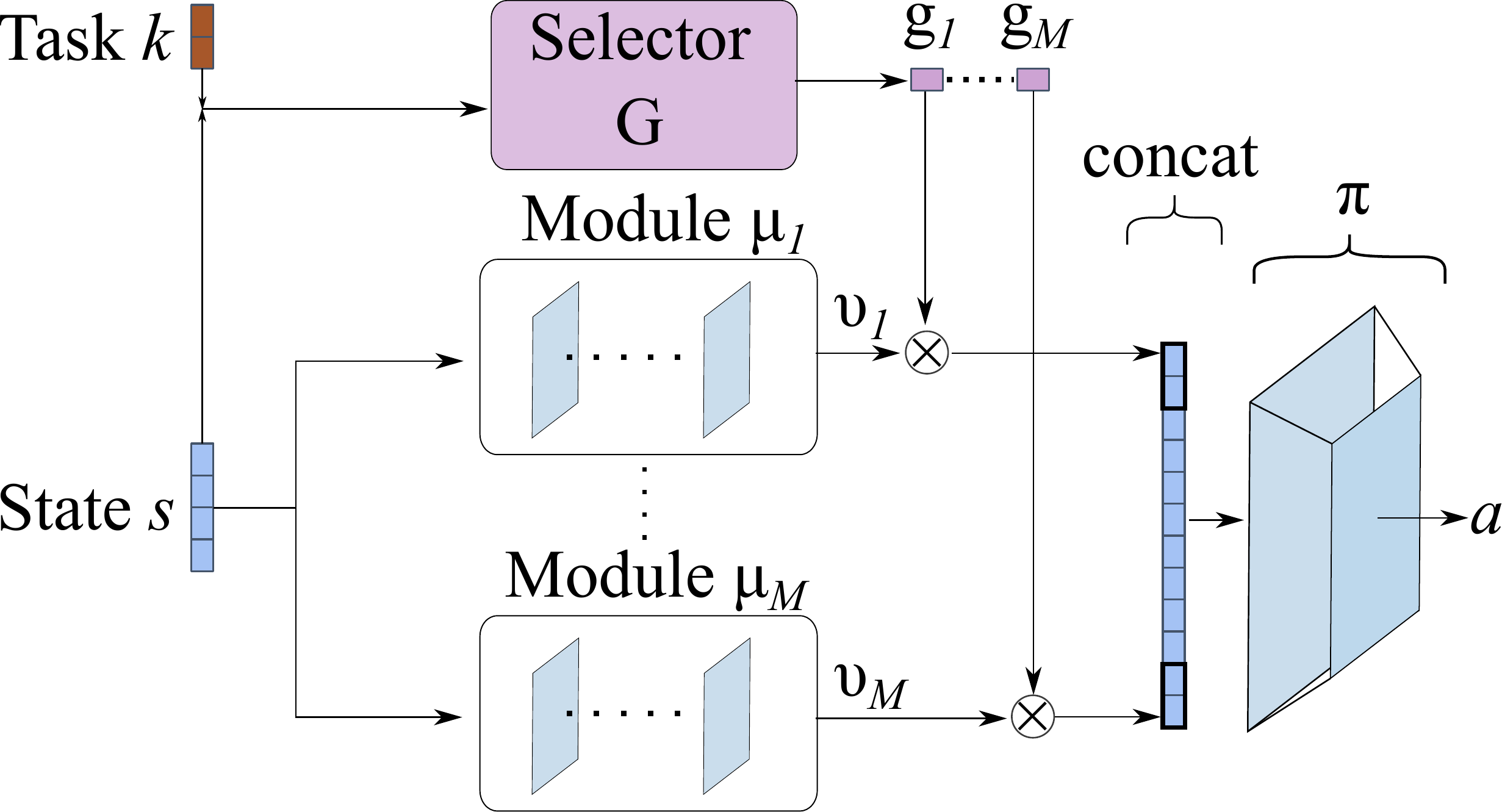}
\caption{MAPS network pipeline. Proto-policy modules operate in parallel using only the state. The selector chooses which to combine based on both state and task.}
\label{fig:arch}
\end{figure}
\par
In our work, we aim to apply the benefits of adaptive architectures and mixture of experts in control problems. Simple policies have been shown to learn short, simple behaviours very well (e.g. reaching an object~\cite{yu2020meta}) but struggle when the tasks become more complicated (e.g. hammering a nail~\cite{Rajeswaran-RSS-18,antotsiou2021adversarial}, performing acrobatics~\cite{2018-TOG-DeepMimic}). Nevertheless, complicated tasks can usually be broken down into shorter, simpler sub-behaviours. Additionally, due to their simple nature, these behaviours are common amongst different tasks. As humans, we can often deduce how tasks can be broken down and the parts that can be shared thanks to our experience, but that can be difficult to model in agents~\cite{liu2016sequential,kirsch2018modular}. Also, similarities might not only be present in the task sequences, but in their general execution as well (e.g. performing a similar task in a different environment, or with a different tool). Our method aims to learn these simpler sub-behaviours by modelling them in the form of modular proto-policies.
\par
The primary contribution of our work is a system that performs MTL by breaking down the tasks into sub-behaviours. To achieve this, we use multiple modules that work in parallel independently. Each of these modules is a proto-policy with the capacity to learn a simple behaviour. We also introduce a way to apply adaptive methods, similar to routing, to control problems. This is done by using a selector mechanism that decides how important each proto-policy is for a particular task, given the current state of the environment. The fact the modules learn only based on the current state and not the task means they learn general behaviours of the state-action space. The selector is the one that forces the proto-policies to learn different behaviours, which are either shared between tasks or used exclusively by one task. This way it achieves positive transfer while minimising negative transfer. The number of proto-policies depends on the number of common or different sub-behaviours present in the tasks, meaning a set of similar tasks can be represented with significantly fewer resources than using a separate agent for each task. A pipeline of our system is shown in Fig.~\ref{fig:arch}. Our system is evaluated on 3 different sets of tasks that exhibit similarities in 3 distinct ways. One has commonalities in its sequence (i.e. same model performs different tasks), one in its environment (i.e. different models perform in the same environment) and one in its actions (i.e. same model performs in different environments). Our system shows better performance compared to a single agent baseline, task-conditioned and multi-headed multi-task methods, as well as state-of-the-art meta-learning. Our method also offers deeper understanding of the learning process by visualising the role of each proto-policy separately. Qualitative results of this investigation can be seen in Section~\ref{sec:bar}.


\section{RELATED WORK}
\textbf{Deep imitation learning} has improved massively over the years~\cite{ho2016generative,finn2017one,duan2017one}. Whilst behavioural cloning (BC)~\cite{pomerleau1991efficient} achieved quick convergence, it required a large dataset and had trouble generalising to unseen states. To alleviate this, \cite{ho2016generative} introduced Generative Adversarial Imitation Learning (GAIL). Although it generalised better, especially with limited datasets, it required interactions with the environment during training. Since then, various works have sought to improve generalisation~\cite{li2017infogail,wang2017robust,hausman2017multi}, and even achieve one-shot learning~\cite{finn2017one,duan2017one}. Similar to our work, ~\cite{hausman2017multi} sought to not only imitate, but also segment skills. It achieved this by learning a multi-modal policy, but unlike ours, it did not take negative transfer into account. Since it was based on GAIL, it also required environmental interaction during training. 
\par
\textbf{Meta-learning} is closely connected to multi-task learning in control, since it can learn a general representation around a set of tasks. In the field of imitation, one such work is Model Agnostic Meta Learning (MAML), presented in~\cite{finn2017model} and optimised in~\cite{antoniou2018train}, where it learnt a common policy for a number of tasks. It then showed this representation point quickly converges to new similar tasks. Whereas this work tried to find one common point between tasks, our work aims to break down tasks into common and uncommon representations. Therefore, it did not accommodate for negative transfer, like ours does. It also required a separate network for every single task. More recently, \cite{singh2020scalable} improved multi-task imitation by utilising the agent's own experience as it trained. Inspired by one-shot learning~\cite{duan2017one}, it encoded the demonstrations into a task embedding and then evaluated the policy roll-outs based on this embedding. If the roll-outs' behaviour was similar to the experts, they were added to the embedding. However, this technique required interaction with the environment during training, which can be demanding in control. In terms of modularity, \cite{alet2018modular} combined meta-learning with modular structures. However, it achieved this by using a set of structures that were dynamically combined to minimise the total loss of tasks. Whereas this can be beneficial in settings that involve a number of layers with potentially different structures, control settings such as the ones in this work have few layers, all of which typically have the same structure.
\par
\textbf{Adaptive networks} have also been used in MTL by adapting their architecture based on the task they are executing. In routing networks, \cite{rosenbaum2017routing} used parallel streams of layers and a router to select which layers to use or skip. While their router was conditioned on each task and input, like our selector, it was applied on every layer sequentially. Our system, on the other hand, trains high level features in parallel and selects them near the end of the pipeline. Additionally, their router was trained using RL, which can be hard to converge, whereas our selector works more like a regulariser that preserves the objectives of MTL and is optimised with gradient descent. Similar to \cite{rosenbaum2017routing}, \cite{sun2019adashare} also connected their modules sequentially with no task-division, but updated their router with gradient descent. Using gradient descent to train the router was also performed in~\cite{ramachandran2018diversity}, which used a number of different types of layers and a router selected the best one based on the task. This is similar to the mixture of experts presented in~\cite{shazeer2017outrageously}. In their work, various different types of layers were selected using a gate and were then linearly combined. Whereas our work also combines features from multiple streams, its main difference from these two works is that it does not use modules that are structurally different, instead it aims to help them learn different high level representations. Using modules and BC to break down a task was also the objective of~\cite{anderson2000behavioral}. But while they managed to divide one task, our work aims to extend upon it and cover multiple tasks. The work that mostly relates to us is~\cite{purushwalkam2019task}, where they used modules both sequentially and in parallel, with a gating mechanism deciding their contribution. Similar to our objective, they wanted the system to learn sub-categories of images but, unlike us, they had prior knowledge of the different sub-categories, and that is hard to obtain in sequential control data.
\par 
\textbf{Multi-task learning in control} has also been achieved using both imitation and reinforcement learning (RL)~\cite{rothfuss2018promp,rakelly2019efficient,sohn2020meta,NEURIPS2020_32cfdce9,ren2020ocean}. However, RL only methods require a reward function for each task rather than expert demonstrations~\cite{sutton2018reinforcement}, as well as interactions with the environment, which are not necessary with many IL methods. Even the RL methods with limited interactions presented in \cite{8944013} required additional knowledge. Contrary to meta-learning, \cite{deisenroth2014multi} tackled MTL in IL and RL by defining the policy as a function of not only the state, but also the task. However, conditioning the entire feature pipeline on the task, as well as using a single fixed policy, can make it hard to adapt to different tasks. This is why we use an adaptive approach that does not condition the features on the task directly. In a similar way, \cite{Xu2018SharedMI} designed a multi-headed MTL system that conditioned only the final features on the task id. Another similarity with our system is that it used many parallel streams of features which are combined at high level. However, all the streams were needed to execute a task, which can be problematic as the number of tasks and streams increases.
\section{PROPOSED METHOD}

{\bfseries Background.} The system is modelled after a Markov Decision Process (MDP). MDPs are expressed with a set of states $\boldsymbol{S}=\left\{s_1,s_2,\dotsc\right\}$, a set of actions $\boldsymbol{A}=\left\{a_1,a_2,\dotsc\right\}$, the probability $\boldsymbol{P}\left(s^\prime | s,a\right)$ that action $a$ will lead to state $s^\prime$ and the reward $\boldsymbol{R}\left(s,a\right)$ for taking action $a$. In control settings the states are the environment and the actions are the robot. In RL, the reward needs to be defined. In IL, it might be available to the experts, but is unknown to the imitation algorithms and needs to be inferred. The goal of imitation is to learn a policy $\pi(a|s)$ i.e. a probability distribution that produces actions similar to the experts. A demonstration dataset ${\mathcal{T}}$ comprises a number of trajectories ${\mathcal{T}}=\left\{\tau_1,\tau_2,\dotsc\right\}$ and each $\tau=\left\{\left(s_1,a_1\right),\left(s_2,a_2\right),\dotsc\right\}$ is a sequence of state-action pairs. 

\subsection{Modular Adaptive Policy Selection (MAPS)}
Our framework, depicted in Fig.~\ref{fig:arch}, consists of $\boldsymbol{M}$ proto-policy modules ${\mu_{i}(
\upsilon |s)}$ that operate in parallel. Similar to single-task imitation networks, each module $\boldsymbol{\mu_i}$ receives $\boldsymbol{s}$, the current state of the environment, but outputs a number of features $\boldsymbol{\upsilon_i}$ instead of a distribution. The selection of the modules is performed by a selector network $G(s,k) \in \mathbb{R}^M$ that outputs $\boldsymbol{M}$ task-relevant scores $(g_1,\dots,g_M)$, one for each module. The scores are then multiplied with the proto-policy features $\boldsymbol{v}$. After that, the $M$ feature vectors are concatenated and inserted to a probability distribution layer $\boldsymbol{\pi}$ that transforms them into a distribution of actions $\boldsymbol{a}$, like a policy network. Therefore, the final actions are
\begin{equation}
    a=\pi^\phi\Big(g_1^w\left(s,k\right)\mu_1^{\theta_1}(s), \dotsc,g_M^w(s,k)\mu_M^{\theta_M}(s)\Big).
\end{equation}
\par
This architecture allows for learning primitives independently from specific tasks. This way, a module does not know which - or if any - tasks use it and how much. This leads to modules that learn commonalities between tasks, but also modules that learn features exclusively tailored to one task. Therefore, the system is capable of performing positive transfer, while avoiding negative transfer.
\par
The two parts of the system, the selector and the proto-policy modules, have very distinct roles. The modules learn separate parts of the task space (be it for many tasks or one) and the selector learns which modules should be used by a task and when. These two objectives form the final loss:
\begin{equation}
    L_{total}=\lambda_{imitate}L_{imitate} + \lambda_{selector}L_{selector}.
    \label{eq:loss}
\end{equation}
The term $L_{imitate}$ is the one that learns the tasks by performing the correct actions. The term $L_{selector}$ is a regulariser that ensures positive transfer is performed, while avoiding negative transfer. For the rest of the paper we will be using behavioural cloning (BC) as the imitation technique for $L_{imitate}$, but any other IL technique (e.g. GAIL) can also be used. Since BC minimises the difference between expert and policy actions, the imitation loss can be written as
\begin{equation}
\begin{gathered}
    L_{imitate}=L_{BC} =\sum_{(s_E,a_E) \in \mathcal{T}_E}\|a-a_E\|^2_2=\\
    \sum_{(s_E,a_E) \in \mathcal{T}_E}\|\pi^\phi\left(g_i^w\left(s_E,k\right)\mu_i^{\theta_i}(s_E), \forall i \in [1,M]\right)-a_E\|_2^2,
\end{gathered}
\label{eq:imi}
\end{equation}
where $(s_E,a_E)$ the state-action pairs of the expert trajectories $T_E$. As evident in~(\ref{eq:imi}), the partial derivatives with respect to $\theta_i$ (modules), $w$ (selector) and $\phi$ (distribution layer) are interconnected. Therefore, training $\theta$ and $w$ jointly allows both of them to benefit from each other. 

\subsection{Module Selection}
Routing networks have often been used in adaptive networks to determine the architecture used at any given point~\cite{rosenbaum2017routing,sun2019adashare}. However, these are usually applied in a sequential fashion, deciding if a layer will be used or skipped based on the task and the features of the previous layer. In our case, the selector is applied once for all the modules, and its decision is about how important each module is for the current state and task. This is done by producing a significance score that is then multiplied with the features of the corresponding module. The final score $g$ of selector $G$ is normalised using softmax:
\begin{equation}
    g(s,k)=softmax\left(G(s,k)\right).
\label{eq:softmax}
\end{equation}
\par
The selector is responsible for deciding which modules to activate, based on the current task $\boldsymbol{k}$ and state $\boldsymbol{s}$. Ideally, this decision would be based on which tasks, or parts of tasks, are similar -- and should share modules -- and which of them are too different and should have dedicated modules for themselves. Unfortunately, such information is not easily obtainable, or even available. To overcome this, our selection mechanism focuses more on the general objective of MTL, rather than specific common characteristics of each task. This objective is broken down into four terms: sharing, exploration, sparsity and smoothness.
\par
{\bfseries Sharing.} The objective of this term is to make different tasks share the same modules for the same input state $\boldsymbol{s}$. Therefore, it represents the positive transfer part of MTL. The concept of sharing has been previously presented in~\cite{sun2019adashare}, where the router encouraged the tasks to share initial layers. We define our term in a similar way, only we have no preference in certain modules and we prefer using the ${L_2}$ loss to increase robustness and offer better balance with the other losses. The final sharing loss is the mean $L_2$ difference between the scores of all the possible pairs of $\boldsymbol{K}$ tasks. For $\boldsymbol{b}$ the total number of samples in $\mathcal{T}_E$, it can be described as
\begin{equation}
    L_{share}=\frac{1}{Mb}\binom{K}{2}\sum_{\subalign{i \in \left[1,M\right],\\s \in \mathcal{T}_E}}\sum_{\subalign{k_1, k_2&=1,\\k_1 &\neq k_2}}^{K}\|g_i(s,k_1)-g_i(s,k_2)\|_2^2.
\label{eq:sharing}
\end{equation}
\par
{\bfseries Exploration.} The objective of this term is to make all modules learn something and avoid mode collapse to one or two modules. This incentivises the system to break the tasks down to multiple -- shareable or not -- elements. The exploration loss chosen for this term is similar to the importance loss presented in~\cite{ramachandran2018diversity} and~\cite{shazeer2017outrageously}, where the coefficient of variation of a batch is minimised. This ensures that in a batch with samples from all tasks, all modules will be used. Therefore, this loss is applied per batch. For $T_k$ as the trajectories of task $k$, and given that $g$ is a $softmax$ distribution and its mean is $\frac{b}{M}$, the exploration loss can be described as
\begin{equation}
        L_{explore}=\frac{M}{b^2}\sum_{i=1}^M\left(\frac{b}{M} - \sum_{k=1}^K\sum_{s\in T_k}g_i(s,k)\right)^2.
\label{eq:exploration}
\end{equation}

{\bfseries Sparsity.} The objective of this term is to ensure each module learns a distinct part of a task. Using the sharing and exploration terms only, it is still possible for all the tasks to use all the modules uniformly. But that renders the selector useless and the modules indistinguishable. The sparsity loss counters this by making each task use only a small subset of modules. Additionally, the combination of sparsity and exploration enables a module to be exclusively used by one task, thus allowing task-specific learning and that minimises negative transfer. Regarding the loss, while~\cite{sun2019adashare} also introduced a sparsity regularisation in the form of log-likelihood, it proved quite unstable in our case. This was due to the logarithmic nature of the loss reaching $-\infty$, which could not be easily balanced with the rest of the terms that minimise at $0$. The Shannon entropy~\cite{shannon1948mathematical}, on the other hand, proved to be a much more stable choice. However, since sparsity wants to avoid uniform distribution at $\frac{1}{M}$, the variable $M$ needs to be incorporated in the loss. For this reason, we shift the entropy function so that its maximum lands on the $\frac{1}{M}$ uniform point. This final sparsity loss is
\begin{equation}
    L_{sparse}=-\frac{1}{Mb}\sum_{i,k=1}^{M,K}\sum_{s\in T_k} g_i(s,k)^{\frac{1}{\log M}}\log g_i(s,k).
\label{maps:eq:sparse}
\end{equation}
\par
{\bfseries Smoothness.} The objective of this term is to provide temporal stability in the selection of modules for each task. This way it avoids erratic module changes between steps. To achieve this, we use the discrete derivative of $g(s,k)$:
\begin{equation}
    L_{smooth}=\frac{1}{Mb}\sum_{i, k=1}^{M,K}\sum_{s\in T_k}\|g_i(s_t,k) - g_i(s_{t-1},k)\|_2^2,
\end{equation}
where $s_t$ and $s_{t-1}$ the states at time steps $t$ and $t-1$ in a trajectory $T_k$.
\par
The total loss of selector $G$ comprises the 4 above-mentioned losses:
\begin{equation}
    L_{selector}=\lambda_1L_{share}+\lambda_2L_{explore} + \lambda_3L_{sparse} +\lambda_4L_{smooth}.
    \label{eq:g}
\end{equation}
It is worth noting that the individual losses compete with each other. Whereas any combination of sharing, exploration and sparsity in pairs makes the loss decrease, the combination of all three makes them compete. This is a manifestation of the opposing MTL objectives of positive transfer without negative transfer. This means careful consideration of the $\lambda$ coefficients is required to achieve balance. For example, if we know there are dissimilar tasks, greater emphasis should be given to exploration (e.g. Gravity set in Section~\ref{sec:result}). On the other hand, if all the tasks are executed in a similar manner, greater emphasis should be given to sparsity and sharing (e.g. MT10 set in Section~\ref{sec:result}). The smoothing loss closely relates to sparsity and is necessary for modular consistency. The combination of sparsity and smoothness ensures that each task consistently uses the same few modules, rather than using a single but different module at every step.

\section{EXPERIMENTS}
\label{sec:exp}
We evaluate the system under 3 different MTL settings covering environmental, model and task variations. We investigate the success rate and stability of the method against a single agent per task baseline, a multi-task agent conditioned on the task id (MT BC), a multi-task agent with multiple heads, one for each task (MT-MH BC), and MAML, which is a standard meta-learning approach for MTL and TL. We also evaluated the method qualitatively by investigating the features each module learns and if the tasks can be broken down into segments, according to our original hypothesis.

\par{\bf{Environmental variation set (HalfCheetah Gravity).}} This set of 5 tasks involves the same agent, performing the same task, only under different environmental conditions. Specifically, they all involve OpenAI's~\cite{1606.01540} HalfCheetah agent running, only under different accelerations of gravity. For $g_r=9.8~m/s$ as the default HalfCheetah's gravity acceleration, the 5 different values used are $0.5g_r$, $0.75g_r$, $g_r$, $1.25g_r$, and $1.5g_r$. These environments were first introduced in~\cite{henderson2017multitask}. To obtain the expert Gravity datasets, an RL agent was trained for each task using TRPO~\cite{schulman2015trust} and the reward function provided in~\cite{henderson2017multitask}. It was then used to generate 35 expert demonstrations.

\par{\bf{Model variation set (HalfCheetah Modified).}} This set of 11 tasks involves performing the same task, under the same environment, only using a different model. Specifically, they all involve OpenAI's HalfCheetah agent running, only the HalfCheetah's body parts are modified. The model has 5 different body parts, and each of them is enlarged or shrunk by 25\% for every task. The body parts include head, torso, thigh, leg, and foot. The original model along with the 5 enlarged and 5 shrunk models make up the 11 tasks of this dataset. Like the Gravity tasks, the Modified environments were introduced in~\cite{henderson2017multitask} and the 35 expert demonstrations were obtained using an RL TRPO agent.

\par{\bf{Task variation set (MT10).}} This set of 10 tasks involves using the same actor performing different tasks in similar environments. Specifically, the sawyer robot gripper is used to perform different tasks manipulating various objects. These tasks include reach, push, pick and place, press button, insert peg in hole, open door, open drawer, close drawer, open window, and close window. These tasks were first introduced in~\cite{yu2020meta}. The expert datasets were obtained in the course of this work using a human expert that manually executed 20 demonstrations of the tasks in the simulated environments.
\par
The selector $G$ has 2 hidden layers, while all the proto-policies $\mu_i$ used have 3. All layers have 128 features each. Both the modules and the selector are optimised using Adam~~\cite{adam} with a learning rate of $3\times10^{-4}$. The batch size for Gravity and Modified is 128, whereas MT10 uses 32. All the baselines use the same policy structure, optimiser, and batch size as the modules. MAML also uses a meta-batch size of 160 for Gravity and Modified, and 40 for MT10. The hyperparameters used are $\lambda_1$=1, $\lambda_2$=0.1, $\lambda_3$=0.5, $\lambda_4$=1 for MT10 and $\lambda_1$=1, $\lambda_2$=0.5, $\lambda_3$=0.001, $\lambda_4$=1 for Gravity and Modified because sparsity was overpowering the BC loss. All sets use $\lambda_{imitate}$=0.75, $\lambda_{selector}$=0.25 and the datasets use 70\% for training and 30\% for validation and tuning. The simulation environment used in this work is Mujoco Pro \cite{kumar2015mujoco}.

\subsection{Quantitative Results}
\label{sec:result}
\begin{figure}[t]
\centering
\includegraphics[width=0.48\textwidth]{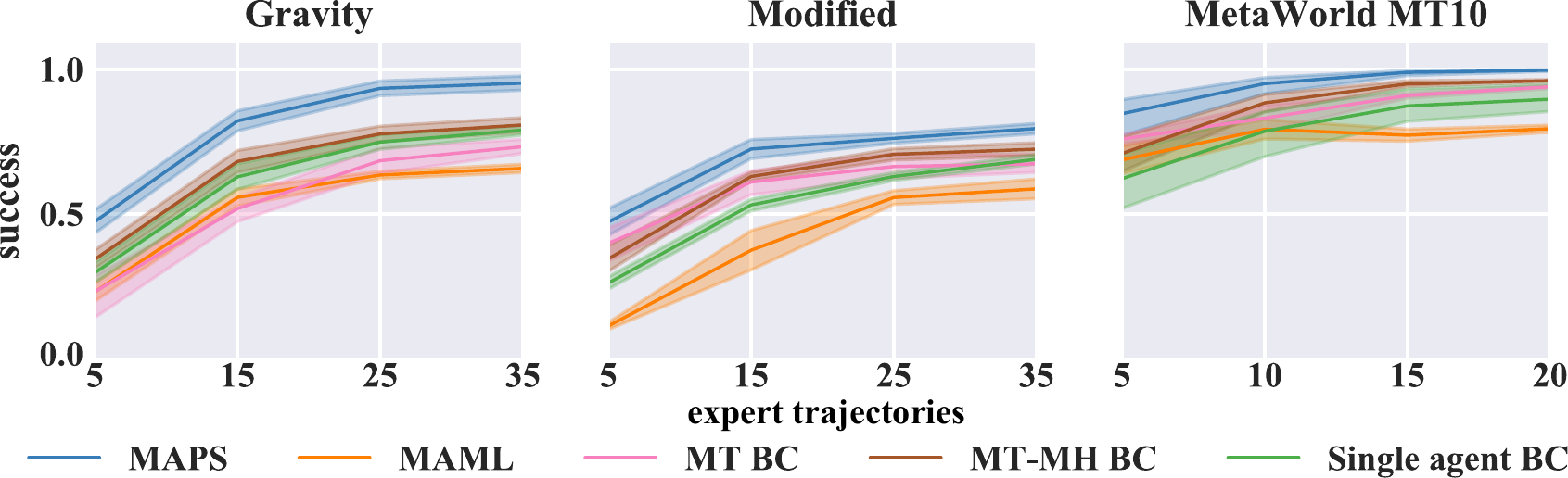}
\caption{Aggregated average results of all the tasks in each of the following sets: Gravity (left), Modified (middle) and MetaWorld's MT10 (right). The runs presented are MAPS with 10 modules, MAML, task-conditioned BC, multi-headed BC and single agent BC over a different number of experts.}
\label{fig:res}
\end{figure}
We evaluate the performance of our method by unrolling the policies on 100 test initial positions for each task, over multiple seeds. We then measure its highest success rate over various different numbers of experts. The results are normalised with the mean reward of the expert trajectories. We then compare the results with 4 different baselines: independent single BC agents for each task, a multi-task BC agent that uses the task id as auxiliary input, a multi-task BC agent that is multi-headed, and agents trained with MAML and then finetuned on each task (up to 10 updates), as described in~\cite{finn2017model}. Whereas MAML can be used in both supervised and RL problems, only its RL version was previously tested in control environments~\cite{finn2017model,yu2020meta}.
\par
\begin{figure}[t]
\centering
\includegraphics[width=0.47\textwidth]{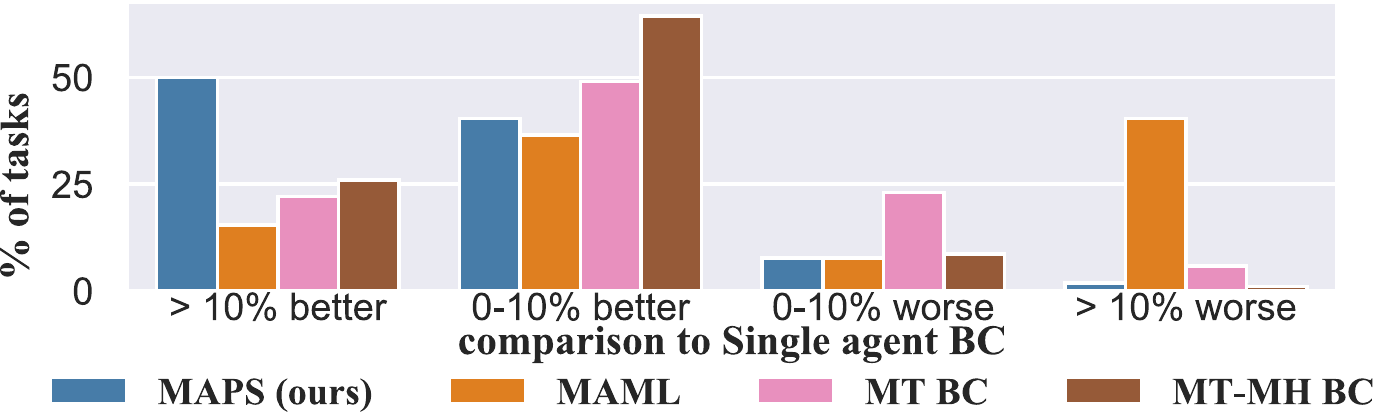}
\caption{Number of tasks each method performed better or worse than Single agent BC, normalised by the total number of tasks.}
\label{fig:hist}
\end{figure}
\begin{figure}[t]
\centering
\includegraphics[width=0.47\textwidth]{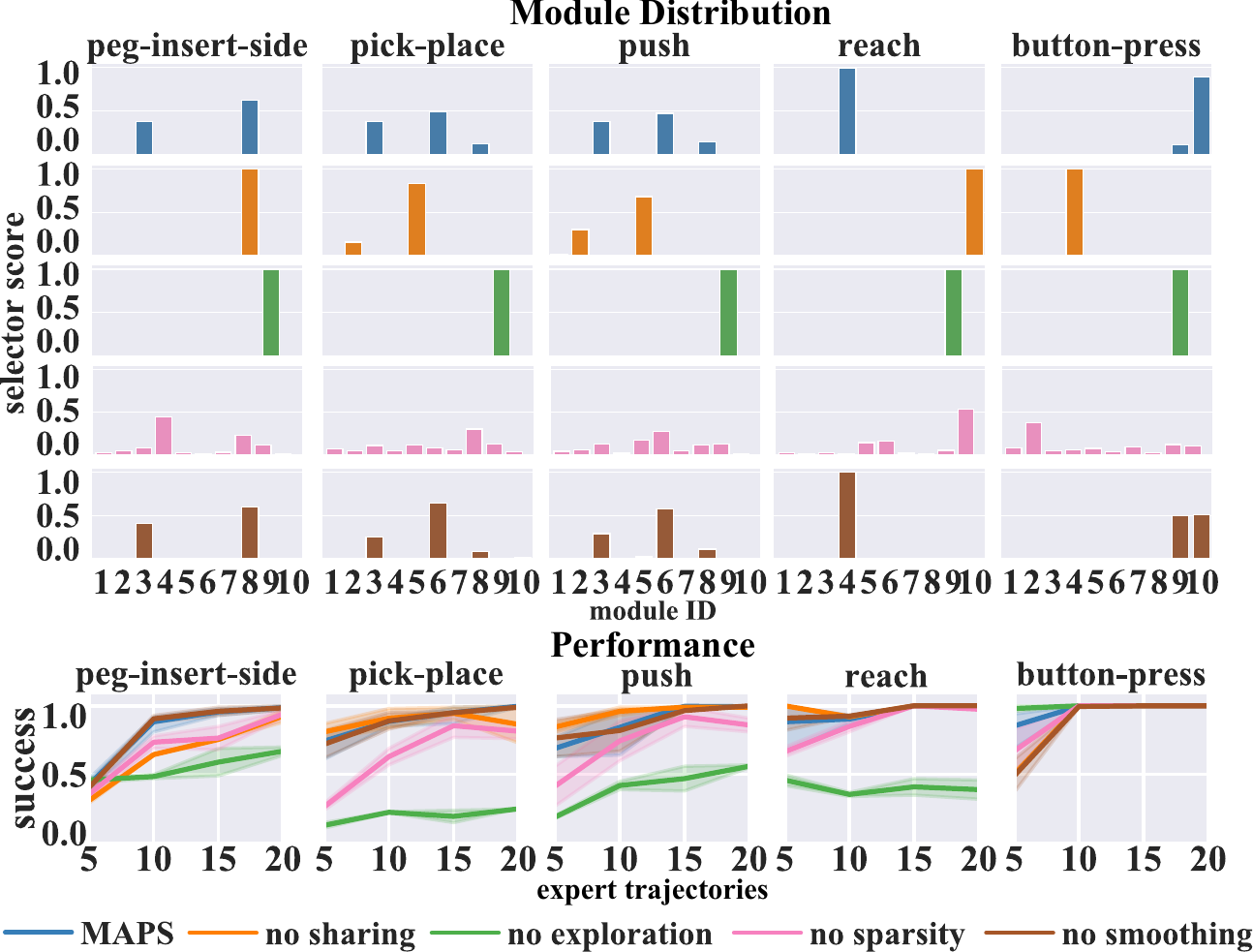}
\caption{Ablations with different selector terms on the MT10 set with 10 modules. {\textbf{Top}}: Bar graphs of the different terms, showing their influence in module selection. {\textbf{Bottom}}: Success rate of the different terms for different experts. It shows the influence of each term on performance.}
\label{fig:abla}
\end{figure}
\begin{figure}[t]
\centering
\includegraphics[width=0.48\textwidth]{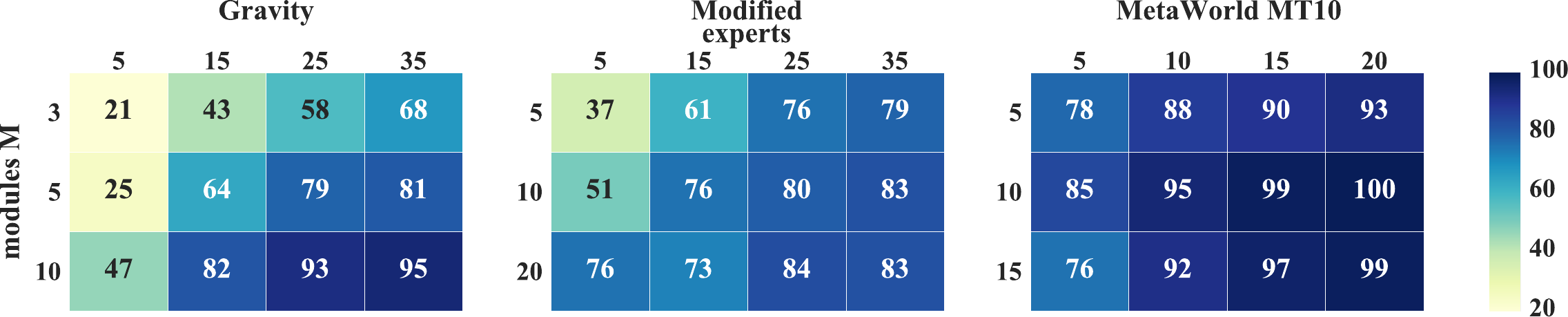}
\caption{Aggregated average success \% of MAPS across all tasks for different number of modules M and experts.}
\label{fig:heat}
\end{figure}
Fig.~\ref{fig:res} shows the experimental results on all the tasks of the 3 sets. As a general rule, MAPS with 10 modules (blue) performs significantly better than all the baselines. Additionally, Fig.~\ref{fig:hist} shows how often the methods perform better or worse compared to the independent BC agent, which has no negative but also no positive transfer. This is done by counting the instances the average performance of a specific task with specific number of experts is better or worse than the single agent BC average performance. This number is then normalised by 104, the total number of different experiment settings (26 tasks under 4 different number of experts). The worst performing one is MAML (orange), which seems to struggle in some tasks, potentially due to them being too different and it being unable to find a single common representation between them. The task-id conditioned MT BC (magenta) performs better than the single agent baseline (green) in most of the tasks, but still suffers from negative transfer in others, as seen in Fig.~\ref{fig:hist}. This is because the baseline relies solely on training to identify the task similarities. Whilst this architecture is more general than MAPS -- and can potentially find similarities of any form -- it can struggle to do so in sets of tasks that are complex or diverse. In these cases, the network can overfit to one or two tasks, which result to negative transfer for the rest. The MT-MH BC baseline (red) performs worse than MAPS by $4-15\%$, but is almost always better than the single agent baseline. This indicates that using both common and task-specific structure is beneficial to multi-task problems. This is seen even more evidently in MAPS, which performs better than the single agent baseline (positive transfer) more than $90\%$ of the time, with $50\%$ of that being a significant (more than $10\%$) improvement. This is mainly thanks to the fact MAPS has a significant amount of structure which allows it to have both shareable and task-specific parts.
\par
Fig.~\ref{fig:abla} details the significance of the various selector terms in the tasks with most variation in the MT10 set using 10 modules. Specifically, it presents the difference in behaviour and performance when one of the terms in removed, compared to MAPS, which uses all of them. The bar graph (top) was produced using only one seed and 10 experts to clearly show the behaviour of each run, whereas the line graphs (bottom) show average performance across multiple seeds and experts. The bar graph shows that by removing sharing, the modules tend to become task-specific, similar to the single agent baseline. By removing exploration, all the tasks end up using only one module, similar to the MT-BC baseline. This is why it is the most important term for the MT10 set in terms of performance. Sparsity, on the other hand, is responsible for making modules task-specific, and its absence also leads to negative transfer. Finally, smoothing has the least impact on performance, since it is there mainly to stabilise sparsity and avoid abrupt module changes.   
\par
Fig.~\ref{fig:heat} shows a heatmap with ablation studies of MAPS with various numbers of modules. The only case where a greater number of modules than tasks was beneficial was in the Gravity set. The other sets' performance seems to plateau at $M=10$. This practically shows how the number of modules should not depend on the number of tasks, but rather the similarities between them. In the MT10 set, where we expect more similarities between the tasks, we see the success of $M=5$ is comparable to the one of $M=10$. Therefore, even with half the size of the single agents' resources, MAPS can still achieve better performance.

\subsection{Qualitative Module Understanding and Task Breakdown}
\label{sec:bar}
\begin{figure}[t]
\centering
\includegraphics[width=0.48\textwidth]{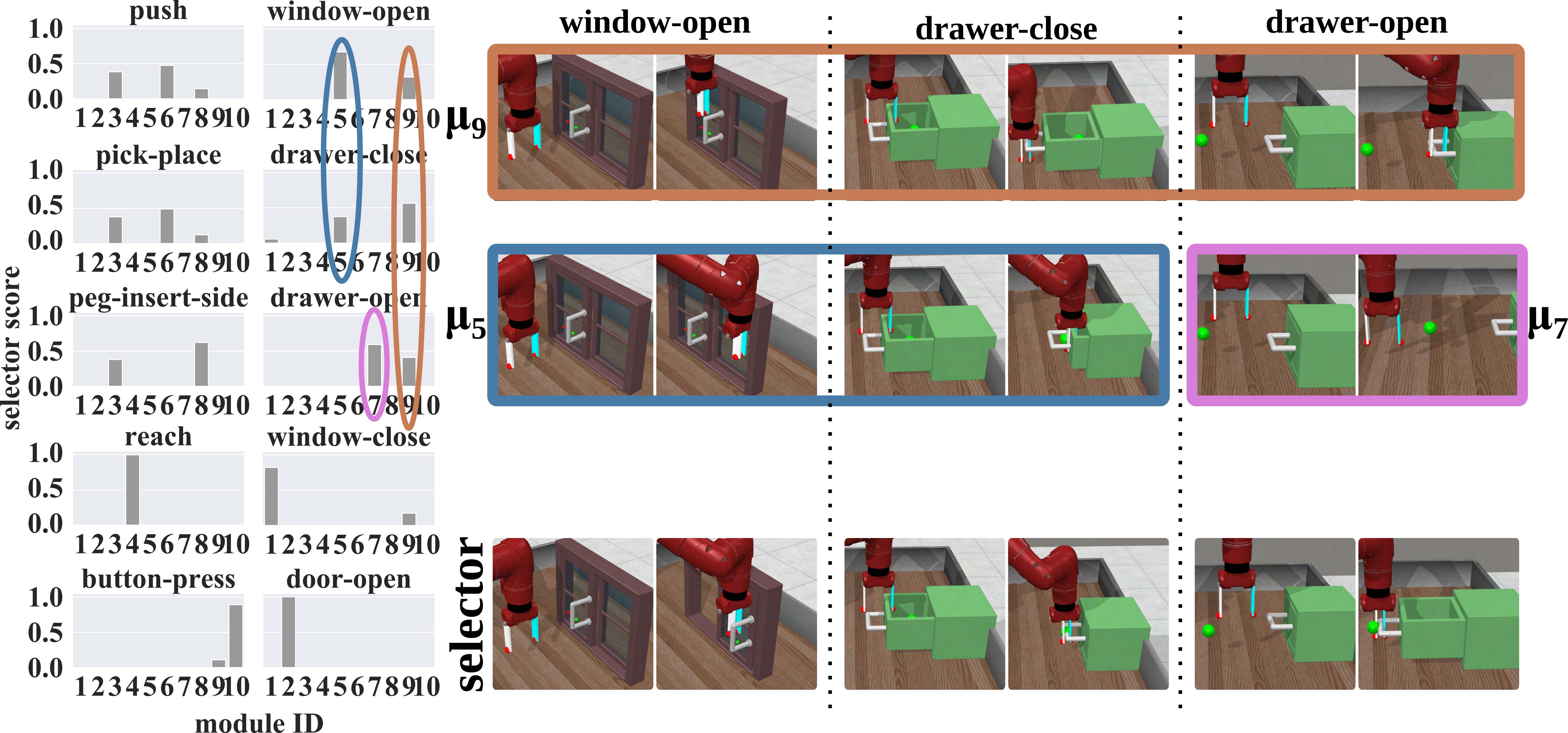}
\caption{Break down of the MT10 tasks with 10 modules (left) and visual examples of individual modules (right). Tasks window-open and drawer-close share ``grab handle'' $\mu_9$ and ``move forward'' $\mu_5$ modules. Task drawer-open shares $\mu_9$ but uses $\mu_7$ as a task-specific ``move backwards'' module$^4$.}
\label{fig:bars}
\end{figure}

We also evaluate how well MAPS manages to break down task behaviour in the MT10 set. Fig.~\ref{fig:bars} (left) presents the occurrence of every module in each individual task. This shows the influence of the selector losses, as well as the common or task-specific nature of the modules. 
\par
The Fig.~\ref{fig:bars} bar graph shows our initial hypothesis is correct and tasks are composed of shareable modules, task-specific modules, or a combination of the two. An example of this is the subset of window-open, drawer-open, and drawer-close tasks. These tasks use both shareable and task-specific modules. The most similar tasks are window-open and drawer-close, which use the same modules $\mu_5$ and $\mu_9$. An example of applying these modules individually on each task can be seen in Fig.~\ref{fig:bars} (right). The figure shows these two modules have learnt two distinct behaviours: $\mu_9$ (orange) has learnt to reach (and grab when possible) the handle, and $\mu_5$ (blue) has learnt to move forward and reach the target. It is very encouraging then that drawer-open also uses $\mu_9$ to perform the ``grab the handle'' sub-behaviour. However, its movement after that needs to be backwards, compared to the forward movement of $\mu_5$. Therefore, this movement is learnt in a different module $\mu_7$ (magenta), which is task-specific for drawer-open. This behaviour demonstrates how MAPS can perform positive transfer, while avoiding negative transfer. Finally, Fig.~\ref{fig:bars} also shows how the combination of all the modules produces a successful trajectory. It should be noted that even though the sub-behaviour ``grab the handle'' is a natural movement for humans, it is there only because it was present in the trajectories performed by the human expert. Only $\mu_5$ in the drawer-close task is enough to succeed, which is more likely what an RL agent would have learnt. However, MAPS is capable of also finding and modelling minute sub-behaviours that make the trajectories look more natural. 
\footnotetext[4]{Video examples at \url{https://sites.google.com/view/maps-mtl}.}

\section{CONCLUSION AND FUTURE WORK}
\label{sec:conclusion}
We present a new method of imitation in multi-task control problems by breaking down the tasks into distinct sub-behaviours. To achieve this, we use an modular architecture that utilises a multitude of independent proto-policy modules and a selector that forces them to cooperate. Our experiments show that this method can be used in various different MTL settings and can not only improve the success rate, but also reduce the resources required, compared to using separate agents. We also show how the proto-policies can be evaluated individually by qualitatively analysing their behaviour. This indicates that MAPS can also offer a better understanding of what is being learnt in every module, something that is not always easy to do in neural networks~\cite{yosinski2015understanding}.
\par
As a future extension of our work, we would like to see it used in transfer learning problems as well, where the already-learnt modules are used on new tasks. Another extension would be to use MAPS with other IL methods like GAIL. Finally, while this work achieves general task decomposition, we would like to see it extended to structured decomposition by combining it with hierarchical RL~\cite{muller2018driving,wulfmeier2019compositional,chang2021modularity}.

\addtolength{\textheight}{-2.5cm}   




\end{document}